\documentclass[sigconf]{acmart}
\usepackage{balance}
\usepackage{threeparttable}
\usepackage{tabularx}
\usepackage{color}
\usepackage{booktabs}
\usepackage{verbatim}
\usepackage{amsopn}
\usepackage{amsmath}
\usepackage{multirow}
\usepackage{colortbl}
\usepackage{makecell}

\usepackage{hyperref}
\hypersetup{
colorlinks=true,
linkcolor=red,
filecolor=magenta,      
urlcolor=blue,
}

\usepackage{amsfonts}
\usepackage{bbm}
\usepackage{bm}

\usepackage{xspace}

\makeatletter
\DeclareRobustCommand\onedot{\futurelet\@let@token\@onedot}
\def\@onedot{\ifx\@let@token.\else.\null\fi\xspace}

\def\eg{\emph{e.g}\onedot} 
\def\ie{\emph{i.e}\onedot} 
 
 \def\vs{\emph{vs}\onedot}

\makeatother
\usepackage{epstopdf}

\AtBeginDocument{%
  }

\copyrightyear{2023}
\acmYear{2023}
\setcopyright{acmlicensed}\acmConference[MM '23] {Proceedings of the 31st ACM International Conference on Multimedia}{October 29--November 3, 2023}{Ottawa, ON, Canada.}
\acmBooktitle{Proceedings of the 31st ACM International Conference on Multimedia (MM '23), October 29--November 3, 2023, Ottawa, ON, Canada}
\acmPrice{15.00}
\acmDOI{10.1145/3581783.3612856}
\acmISBN{979-8-4007-0108-5/23/10}

\begin{document}

\title{Data Augmentation for Human Behavior Analysis in Multi-Person Conversations} 


\author{Kun Li}
\email{kunli.hfut@gmail.com}
\affiliation{%
    \institution{
    Hefei University of Technology}
    \city{Hefei}
    \state{Anhui}
    \country{China}
    \postcode{230601}
}
\author{Dan Guo}
\email{guodan@hfut.edu.cn}
\authornote{Corresponding authors}
\affiliation{%
    \institution{
    Hefei University of Technology \\ Institute of Artificial Intelligence,
Hefei Comprehensive National
Science Center}
    \city{Hefei}
    \state{Anhui}
    \country{China}
    \postcode{230601}
}
\author{Guoliang Chen}
\email{gouliangchen.hfut@gmail.com}
\affiliation{%
    \institution{
    Hefei University of Technology}
    \city{Hefei}
    \state{Anhui}
    \country{China}
    \postcode{230601}
}
\author{Feiyang Liu}
\email{mffff0120@gmail.com}
\affiliation{%
    \institution{
    Hefei University of Technology}
    \city{Hefei}
    \state{Anhui}
    \country{China}
    \postcode{230601}
}
\author{Meng Wang}
\authornotemark[1]
\email{eric.mengwang@gmail.com}
\affiliation{%
    \institution{
    Hefei University of Technology \\ Institute of Artificial Intelligence,
Hefei Comprehensive National
Science Center}
    \city{Hefei}
    \state{Anhui}
    \country{China}
    \postcode{230601}
}

\renewcommand{\shortauthors}{Kun Li, DanGuo, Guoliang Chen, Feiyang Liu,\& Meng Wang}

\begin{abstract} 
In this paper, we present the solution of our team HFUT-VUT for the MultiMediate Grand Challenge 2023 at ACM Multimedia 2023.  
The solution covers three sub-challenges: bodily behavior recognition, eye contact detection, and next speaker prediction. We select Swin Transformer as the baseline and exploit data augmentation strategies to address the above three tasks. 
Specifically, we crop the raw video to remove the noise from other parts. At the same time, we utilize data augmentation to improve the generalization of the model. As a result, our solution achieves the best results of 0.6262 for bodily behavior recognition in terms of mean average precision and the accuracy of 0.7771 for eye contact detection on the corresponding test set. In addition, our approach also achieves comparable results of 0.5281 for the next speaker prediction in terms of unweighted average recall. 
\end{abstract}



\maketitle

\begin{figure}[ht!]
\centering
\includegraphics[width=1.0\linewidth]{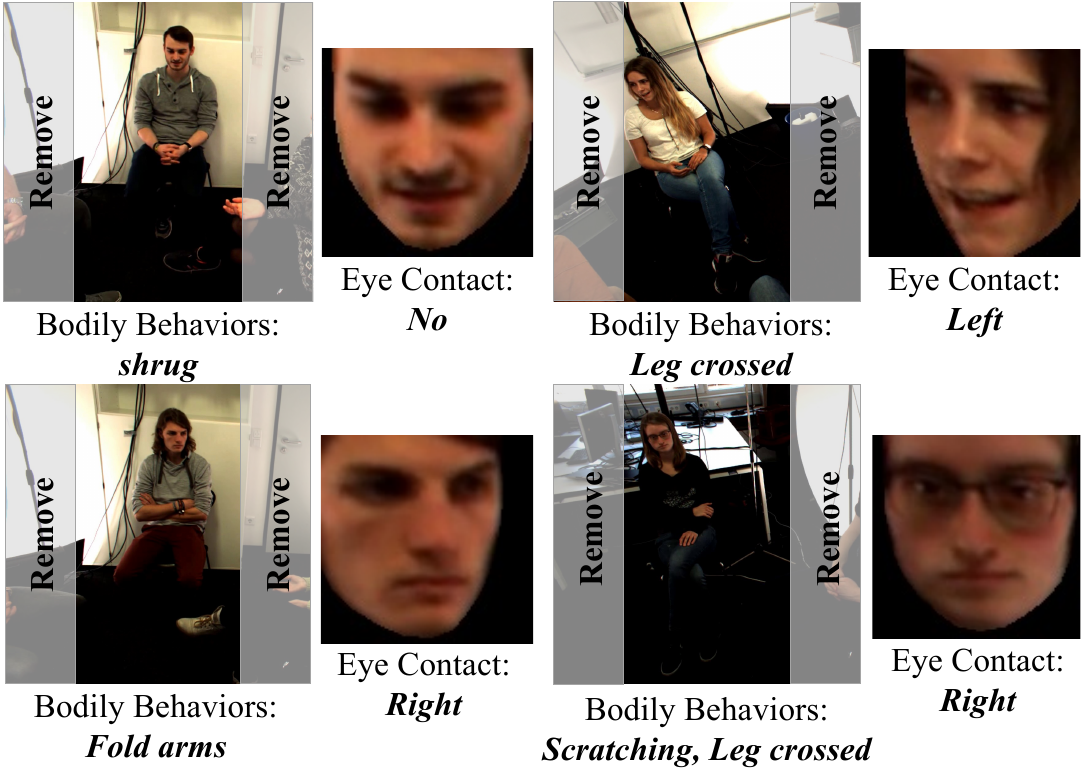}
\vspace{-2.7em}
\caption{Video samples from the MPIIGroupInteraction~\cite{muller2018detecting} and BBSI~\cite{balazia2022bodily} datasets. 
For bodily behavior recognition, we remove the irrelevant background. For Eye contact detection, we use OpenFace~\cite{baltrusaitis2018openface} to extract face region. For the next speaker prediction, we use the raw frame as input data. 
}
\vspace{-2.0em}
\label{fig:feature}
\end{figure}

\section{Introduction}
\begin{figure*}[ht!]
\centering
\includegraphics[width=1.0\linewidth]{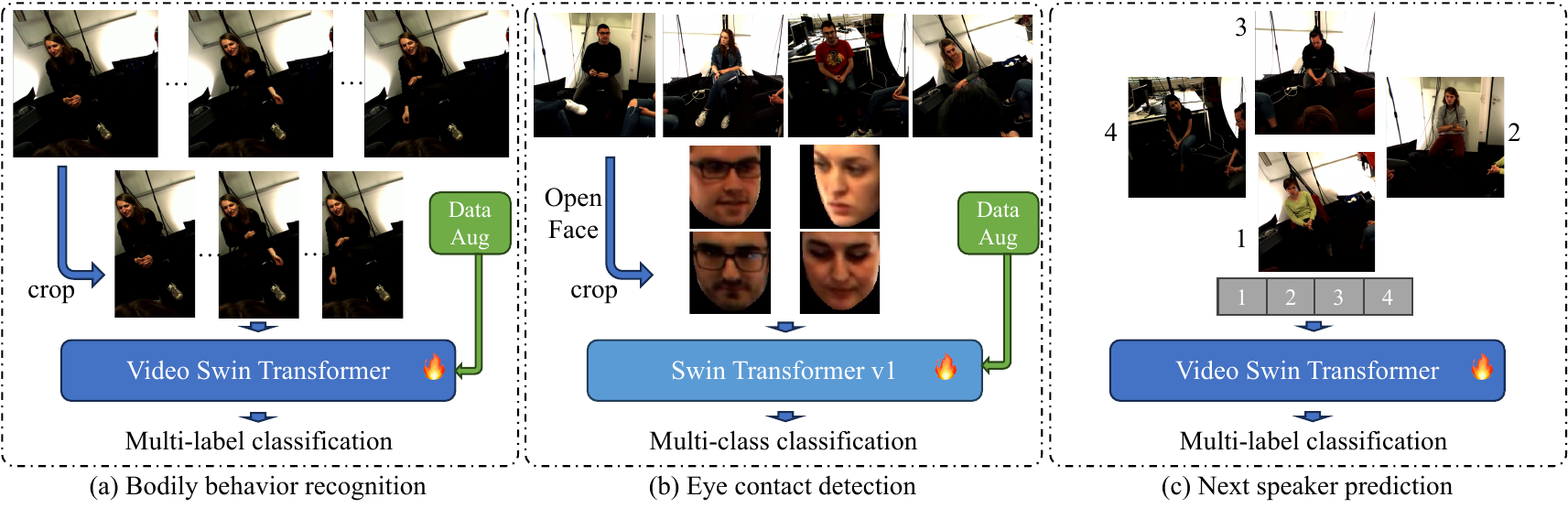}
\vspace{-2.8em}
\caption{The architecture overview of our solution for (a) bodily behavior recognition, (b) eye contact detection, and (c) next speaker prediction. 
For task a, we remove the irrelevant background for action classification. For task b, we extract the face region to train the model. For task c, we directly use the raw frames as input data for the model. 
}
\vspace{-1.5em}
\label{fig:method}
\end{figure*}
Multi-modal Group Behaviour Analysis for Artificial Mediation (Multimediate) is a grand challenge at ACM Multimedia 2023. It is launched based on the MPIIGroupInteraction dataset~\cite{muller2018detecting}  and the Noxi datasets~\cite{cafaro2017noxi} and requires researchers to study key issues in group behavior perception and analysis. This year, the challenge included six tasks, \ie, bodily behavior recognition, eye contact detection, next speaker prediction, backchannel detection, engagement estimation, and agreement estimation. 
We only participated in the first three tasks. 
The bodily behavior recognition~\cite{muller2023multimediate} is formulated as a 14-class multi-label classification task, and participants are required to predict the likelihood of each behavior class by the 64-frame video snippets. 
The eye contact detection task~\cite{muller2018robust} requires predicting the absolute eye contact classes after the end of the 10s context window. 
The next speaker detection task~\cite{muller2021multimediate} requires participants to predict the speaking status of each participant one second after the end of the 10s context window. 
Our main contributions are summarized as follows:
\begin{itemize}
\item We explore different data augmentation strategies while training the model, directly leading to accuracy improvements on the baseline model. 
\item We explore data processing strategy in the bodily behavior recognition task by filtering irrelevant background information to improve the model's classification accuracy. 
\item For both the bodily behavior recognition and eye contact detection tasks, our solution achieves the top-1 performance on the leader board with the mAP of 0.6262 and accuracy of 0.7771, respectively. 
\end{itemize}

\section{Methodology}
\subsection{Bodily Behaviour Recognition}
Bodily behavior recognition aims to recognize the category of behavior (\eg, \textit{shrug}, \textit{leg crossed}, and \textit{fold arms} in Figure~\ref{fig:feature}) through the given videos from different views, so it is formulated as a multi-label classification task. 
As shown in Figure~\ref{fig:method}, given the raw video sequence $V\in \mathbb{R}^{T\times 1000\times 1000\times 3}$, we first remove the irrelevant background to construct the cropped video $V^{\prime}\in \mathbb{R}^{T\times 1000\times 600\times 3}$. 
Subsequently, following the official baseline~\cite{muller2023multimediate}, we utilize the video swin transformer~\cite{liu2022video} as the baseline to build a multi-label classification model. 
Data augmentation plays a crucial role in deep learning. By expanding and diversifying the training dataset, it enhances the model's ability to generalize and learn robust representations. As shown in Figure~\ref{fig:method}, we can see that the participants' clothing and background colors in the video are very similar, which can easily cause confusion between the foreground and background, affecting the accuracy of model recognition. Based on this characteristic, we utilize the ``ColorJitter'' data augmentation strategy to enhance the sensitivity of the model to the background. For detailed analysis and experimental results, please refer to Section~\ref{sec:re_bbr}. 

\subsection{Eye Contact Detection} 
Eye contact detection aims to estimate the gaze direction of each participant. In this sub-challenge, eye contact is defined as a discrete indication of whether a participant is looking at another participant's face. 
Thus, this sub-challenge can be formulated as a multi-class (\ie, left, frontal, right, and no eye contact) classification task.  
Following previous work~\cite{muller2021multimediate,ma2022ta}, as shown in Figure~\ref{fig:method} (b), we first use the OpenFace 2.0~\cite{baltrusaitis2018openface} to detect and align faces from the original video and crop them to 224$\times$224 images. Then, these images are input to swin transformer~\cite{liu2021swin} for multi-class classification. 
Similar to the bodily behavior recognition task, we utilize the ``ColorJitter'' and ``Lighting'' data augmentation strategies into the training process to improve the robustness of the model. 

\subsection{Next Speaker Prediction}
Next speaker prediction aims to predict the speaking status of each participant one second after the end of the 10s context window. Thus, this sub-challenge can be formulated as a multi-label classification task. 
Unlike previous work~\cite{ma2022ta}, we attempt to predict the next speaker directly from the original video frame sequence without using cropped facial features. 
As shown in Figure~\ref{fig:method} (c), we first extract the video frames, then concatenate these frames according to the order in the annotation file. Subsequently, the concatenated frames are input to the video swin transformer~\cite{liu2022video} to perform multi-label classification to predict the next speaker. 
Due to the limited time available for the challenge, we did not have time to attempt to extract facial features or apply data augmentation to improve the accuracy of predictions.

\section{Experiments}
\subsection{Experiment Setup}
\subsubsection{Datasets}
The MultiMediate'23 Challenge utilized the MPIIGroupInteraction dataset~\cite{muller2018detecting} with the BBSI annotation~\cite{balazia2022bodily} for bodily behavior recognition, eye contact detection, and next speaker prediction. 
The dataset contains bodily behavior categories and one background category, with a total duration of 26 hours. 
For bodily behavior recognition task, there are a total of 43,712 video clips, of which the number of video instances in the training, validation, and test sets are 31,221, 11,496, and 995, respectively. 
For eye contact detection and next speaker prediction sub-challenges\sloppy, the data were taken from the MPIIGroupInteraction dataset, consisting of 22 group conversations and 78 German-speaking participants. 
Each group consists of three to four participants, with a video camera behind each participant to record the content of the current participant's perspective. 
Specifically, for the eye contact detection task, there are 4,504 training samples, 1,672 validation samples, and 1,848 test samples.
For the next speaker prediction task, there are 7,546 training samples, 4,036 validation samples, and 5,222 test samples. In addition, this challenge also provided the individual viewpoints of each participant, so each sample has 3 to 4 viewpoints. The training sample has 30,184 multi-view videos, while the validation and test samples consist of 16,144 and 20,888 multi-view video files, respectively. 

\subsubsection{Implementation Details}
\textbf{Data Processing:} For the bodily behavior recognition task, the resolution of the officially provided raw video was 1000$\times$1000. However, due to the camera viewpoint, the rest of the participants occasionally appeared in the video, which caused some interference to the action recognition of the participants in the middle position. 
Therefore, we made an initial crop of the video by cropping the left and right by 200 pixels each to get a video with the resolution of 600$\times$1000. 
For the eye contact detection task, we first utilized OpenFace~\cite{baltrusaitis2018openface} to detect and align faces from the original video and cropped them to 224$\times$224 images. 
For frames where faces could not be detected, we used a black-filled image instead.
For the next speaker prediction task, we only adopted the last frame of each video. 
Considering that the group interaction has 4 different viewpoints, we use the order of viewpoint labeling in the annotation to combine the frames into a sequence to feed into the model. 

\noindent \textbf{Model Parameters:} For the video swin transformer~\cite{liu2022video} in bodily behavior recognition, we use the base version pre-trained on kinetics-400 dataset~\cite{kay2017kinetics}.   
We set the batch size to 2 with an initial learning rate of 3e-4. 
In addition, the model is trained in 10 epochs with an early stop strategy. 
For bodily behavior recognition and next speaker prediction tasks, our code is implemented with the open-source toolbox MMAction2~\cite{2020mmaction2}. 
For eye contact detection, we implement the code with the toolbox MMPreTrain~\cite{2023mmpretrain}. 

\noindent \textbf{Evaluation Metrics:}
The bodily behavior recognition is a multi-label classification task, the mean average precision (mAP) is adopted as the evaluation metric. 
Eye contact detection is formulated as sing-label classification; thus, we utilize the accuracy to evaluate the model's performance. 
For the next speaker prediction, we adopt unweighted average recall (UAR) over all samples to evaluate the model. 

\begin{table}[t]
\caption{Performance comparison of different methods on bodily behavior recognition in terms of mAP (mean average precision) metric.}
\vspace{-1.0em}
\resizebox{1.0\linewidth}{!}{
\begin{tabular}{c|c|cc}
\toprule
Setting &  Method   & Val & Test \\ \hline
random & Baseline~\cite{muller2023multimediate} & 0.0884 & 0.2355 \\ 
w/o bg, frontal view & Baseline~\cite{muller2023multimediate} & 0.3974      & 0.5315 \\
w/o bg, mean of all views & Baseline~\cite{muller2023multimediate} & 0.4099 & 0.5628 \\ \hline 
w/o bg, frontal view & PoseC3D~\cite{duan2022revisiting} & 0.3810 & - \\
w/o bg, cropped frontal view & 
\textbf{Ours} & \textbf{0.4822} & \textbf{0.6262} \\
\bottomrule
\end{tabular}}
\vspace{-1.0em}
\label{tab:human}
\end{table}

\begin{table}[t]
\caption{Ablation study of data augmentation in bodily behavior recognition in terms of mAP metric. ``CJ'' and ``RE" denote the ``ColorJitter'' and ``RandomErasing'' data augmentation strategies, respectively.}
\vspace{-1.0em}
\resizebox{1.0\linewidth}{!}{
\begin{tabular}{c|c|cc}
\toprule
Setting       & Method & Val & Test \\ \hline
w/o bg, frontal view  & Baseline ~\cite{muller2023multimediate}          &  0.4679  & - \\
w/o bg, frontal view     & ~\cite{muller2023multimediate}+ CJ + RE &  0.4688 & 0.6143     \\ \hline
w/o bg, cropped frontal view & Baseline~\cite{muller2023multimediate}          &  0.4757  & 0.6139     \\
w/o bg, cropped frontal view &  ~\cite{muller2023multimediate}+CJ+RE (\textbf{Ours}) & \textbf{0.4822} & \textbf{0.6262}    \\ \bottomrule
\end{tabular}}
\vspace{-1.5em}
\label{tab:abl_bodily}
\end{table}

\subsection{Results for Bodily Behavior Recognition}
\label{sec:re_bbr}
As shown in Table~\ref{tab:human}, we report the performance comparison between our approach and other methods. 
Considering the large amount of data for behavior recognition, we first attempt to use the officially provided skeleton data. Specifically, we trained the PoseC3D model~\cite{duan2022revisiting} using frontal view skeleton data as input and achieved a mAP of 0.3810 on the validation dataset. 
We believe that directly using the PoseC3D model cannot achieve good results in action recognition, and further adjustments are needed to the input features. 
Considering the limited number of online evaluations, we did not make in-depth improvements on the skeleton-based action recognition models. 
Compared with the baseline with the frontal view as input, our approach achieves the best performance of 0.6262 on the test set with 9.47\% improvements. 
Due to limitations in the training duration of the model, we did not use videos from three views to train the model. 
Observing the third row of Table~\ref{tab:human}, we can see that model using videos from all views can significantly improve performance (\ie, 0.5628 \vs 0.5315 on the test set).    
We believe our method will further improve if we use videos from three views to train the model. 
As shown in Table~\ref{tab:abl_bodily}, we give the ablation study results of our method. 
The RandomErasing~\cite{zhong2020random} strategy randomly erases a rectangle region with random values, which is widely used in image and video classification tasks.  
Specifically, we increase the probability of ``RandomErasing'' from the initial 0.25 to 0.5.
Observing the first and third rows of Table~\ref{tab:abl_bodily}, we can see that the cropped videos can improve the mAP, \ie, (0.4757 \vs 0.4679 on the val set). Observing the third and fourth rows, we can conclude that the ``CorlorJitter'' and ``RandomErasing'' data augmentation strategies can further enhance the mAP of the model (\ie, 0.6262 \vs 0.6143 on the test set). 

\subsection{Results for Eye Contact Detection}
The validation and test set results of eye contact detection are shown in Table~\ref{tab:eye}. Compared with SVM-RBF~\cite{muller2021multimediate}, our method achieves a relative improvement of 49.44\% in terms of accuracy on the test set. MH~\cite{fu2021using} and Baseline~\cite{muller2022multimediate} employ different feature sampling strategies. For improved accuracy, the baseline trains four eye contact detection models for each seating position and produces an accuracy of 0.64 and 0.576 on the validation and test sets, respectively.
We use the face image extracted by OpenFace 2.0~\cite{baltrusaitis2018openface} as input. Compared with the latest methods (\ie, TA-CNN~\cite{ma2022ta} and TA-CNN~\cite{ma2022ta}$^\dagger$), our method achieves the best accuracy of 0.8122 on validation set and 0.7771 on the test set. 
The significant improvement means that our method has superior generalization capabilities. 

The ablation study results of eye contact detection are shown in Table~\ref{tab:abl_eye}. Firstly, we use the last frame face feature of each video as input and use Swin V1-B (base model) and Swin V1-L (large model) as baseline methods. 
On the validation set, Swin V1-B achieves 0.8175, \ie, 0.71\% higher than Swin V1-L. 
We attribute this to the fact that the Swin V1-L model may be difficult to fully converge on this small dataset. 
Secondly, we expand the dataset using face images from the last 5 frames of each video and add two data augmentation methods. 
``ColorJitter'' can randomly change the brightness, contrast, and saturation of an image. ``Lighting'' can adjust image lighting using AlexNet-style PCA jitter. 
These two data augmentation strategies are widely image transforms. 
Both data augmentation strategies contribute enhancement to the results on the validation set. 
Finally, we experiment on the latest Swin V2 model~\cite{liu2022swin} as well and utilize the data augmentation methods mentioned above, reaching 0.8307 on the validation set. 
Due to limitations on the number of tests, we are unable to provide results for the test set. 

\begin{table}[t]
\caption{Performance comparison of different methods on eye contact detection in terms of accuracy metric. $^\dagger$ denotes that both training and validation sets are used to train the model.}
\vspace{-1.0em}
\resizebox{1.0\linewidth}{!}{
\begin{tabular}{c|c|c|cc}
\toprule
Feature & Method & Venue  & Val & Test \\ \hline
OpenFace & SVM-RBF~\cite{muller2021multimediate} & MM'21 & & 0.52 \\
MHI & MH~\cite{fu2021using} & MM'21 & - & 0.56  \\ 
Gaze   & Baseline~\cite{muller2022multimediate} & MM'22 & 0.61 & - \\
Headpose& Baseline~\cite{muller2022multimediate} & MM'22 & 0.61 & - \\
Gaze + Headpose & Baseline~\cite{muller2022multimediate} & MM'22 & 0.64 & 0.576 \\
OpenFace  & TA-CNN~\cite{ma2022ta} & MM'22 & 0.7605    & 0.7148 \\
OpenFace  & TA-CNN~\cite{ma2022ta}$^\dagger$ & MM'22 &  - & 0.7261 \\ \hline

OpenFace & \textbf{Ours}  & 2023 & \textbf{0.8122} & \textbf{0.7771} \\ \hline     
\end{tabular}}
\vspace{-1.5em}
\label{tab:eye}
\end{table}

\begin{table}[]
\caption{Ablation study of our model in eye contact detection in terms of Acc metric. \#1 and \#5 denote 1 and 5 frames are used, respectively. ``CJ'', ``L" and ``RA'' denote the ``ColorJitter'', ``Lighting'' and ``RandAugment'' data augmentation strategies, respectively. }
\vspace{-1.0em}
\begin{tabular}{c|c|c|c}
\toprule
Feature          & Backbone & Method & Val \\ \hline
OpenFace (\#1) & Swin V1-B~\cite{liu2021swin}  & Baseline & 0.8175 \\
OpenFace (\#1) & Swin V1-L~\cite{liu2021swin}  & Baseline & 0.8104  \\
\hline
OpenFace (\#5) & Swin V1-B~\cite{liu2021swin}  & Baseline & 0.8122  \\
OpenFace (\#5) & Swin V1-B~\cite{liu2021swin}  & +CJ & 0.8188 \\
OpenFace (\#5) & Swin V1-B~\cite{liu2021swin}  & +L & 0.8205 \\
\hline
OpenFace (\#1) & Swin V2-B~\cite{liu2021swin}  & Baseline & 0.8272  \\
OpenFace (\#1) & Swin V2-B~\cite{liu2021swin}  & +RA & 0.8277  \\
OpenFace (\#1) & Swin V2-B~\cite{liu2021swin}  & +L & \textbf{0.8307}  \\ \bottomrule
\end{tabular}
\vspace{-1.5em}
\label{tab:abl_eye}
\end{table}

\subsection{Results for Next Speaker Prediction}
As shown in Table~\ref{tab:next}, we report the experimental results compared to previous methods. The baseline methods utilized static and dynamic features extracted over the complete input video (frame-by-frame) via OpenFace 2.0. 
GLFVA~\cite{birmingham2021group} used the combination of group-level focus on visual attention features and publicly available audio-video synchronizer models, which achieved the best score of 0.632 on the test set in this sub-challenge. 
TA-CNN~\cite{ma2022ta} utilized group-level features, achieving 0.6538 and 0.5766 on the validation and test sets, respectively. 
Similar to the sampling strategy of TA-CNN, we extracted the last frame of the video from different viewpoints for each sample. Then, the frame sequence organized according to the order of viewpoints is used as input to the network. 
Considering the limited computing resource and evaluation chance, we have not yet had time to experiment with the methodology of previous work~\cite{birmingham2021group,ma2022ta} extracting facial features as input to the model.
Unlike the above methods with individually cropped face images, we attempted to use the complete original frames to capture information about changes in the physical behavior of different participants and predict the next speaker based on it. Although our results on the test set reached only 0.5281, a new attempt was made to explore the task of predicting the next speaker. 

\begin{table}
\caption{Performance comparison of different methods on Next Speaker Prediction in terms of UAR (unweighted average recall) metric. $^\dagger$ denotes that both training and validation sets are used to train the model.}
\vspace{-1.0em}
\begin{tabular}{c|c|c|cc}
\toprule
Feature       & Method  & Venue & Val & Test \\ \hline
Group S       & Baseline~\cite{muller2021multimediate} & MM'21&0.60      & - \\
Group D       & Baseline~\cite{muller2021multimediate} & MM'21&0.53      & - \\
Group (S+D)   & Baseline~\cite{muller2021multimediate} & MM'21&0.60      & - \\
Video \& Audio & GLFVA~\cite{birmingham2021group} & MM'21 & - & \textbf{0.632} \\
OpenFace & TA-CNN~\cite{ma2022ta} & MM'22 & 0.6538    & 0.5766     \\
OpenFace & TA-CNN~\cite{ma2022ta}$^\dagger$ & MM'22 &- & 0.5965    \\ 
\hline
Raw frame & \textbf{Ours} & 2023 & - & 0.5281 \\
\bottomrule
\end{tabular}
\vspace{-1.5em}
\label{tab:next}
\end{table}

\section{Conclusions}
In this work, we present our solution developed for the Multimediate grand challenge hosted at ACM Multimedia 2023. 
For bodily behavior recognition, by incorporating the video swin transformer with heuristic data processing and data augmentation strategies, our approach achieved first place with the mAP value of 0.6262. 
For eye contact detection, the face region extracted by OpenFace from the last five frames is input to the Swin transformer baseline, resulting in first place with an accuracy of 0.7771. 
For the next speaker prediction task, we utilized raw frames as the input of the video swin transformer and achieved a comparable performance of 0.5281 on the test set. 

In the future, we would like to apply skeleton-based action recognition methods~\cite{li2023joint} and mask irrelevant backgrounds with segmentation map~\cite{agrawal2023multimodal} for bodily behavior recognition. 
In addition, we also would like to exploit the temporal relation~\cite{li2021proposal,guo2019connectionist} in the video sequence for eye contact detection. 
For the next speaker prediction task, we plan to use the pre-trained model on the face recognition dataset to extract the face region feature to improve the classification accuracy of the model.

\begin{acks}
This work was supported by the National Key R\&D Program of China (2022YFB4500600), the National Natural Science Foundation of China (62272144, 72188101, 62020106007, and U20A20183), and the Major Project of Anhui Province (202203a05020011). 
\end{acks}

\bibliographystyle{ACM-Reference-Format}
\balance
\bibliography{sample-base}

\end{document}